%% file: conference_101719.tex
\documentclass[conference,comsoc]{IEEEtran}
\IEEEoverridecommandlockouts
\usepackage{amsmath,amssymb,amsfonts}
\usepackage{algorithmic}
\usepackage{graphicx}
\usepackage{textcomp}
\usepackage{xcolor}
\usepackage{url}
\usepackage[hidelinks]{hyperref}
\usepackage[nolist,nohyperlinks]{acronym}
\usepackage[compress]{cite}
\usepackage{layouts}
\usepackage{booktabs}
\usepackage{etoolbox}
\usepackage{multirow}
\ifCLASSOPTIONcompsoc
    \usepackage[caption=false, font=normalsize, labelfont=sf, textfont=sf]{subfig}
\else
\usepackage[caption=false, font=footnotesize]{subfig}
\fi
\makeatletter
\patchcmd{\@makecaption}
  {\scshape}
  {}
  {}
  {}
\makeatother
\input{acronyms}
\input{commands}

\begin{document}
\bstctlcite{BSTcontrol}

\title{\textsc{Flexible}: \textbf{F}orecasting Cellular Traffic by \textbf{L}everaging \textbf{Ex}plicit \textbf{I}nductive Graph-\textbf{B}ased \textbf{Le}arning}
\author{
    \IEEEauthorblockN{
        Duc-Thinh Ngo\IEEEauthorrefmark{1}\IEEEauthorrefmark{2}, Kandaraj Piamrat\IEEEauthorrefmark{2}, Ons Aouedi\IEEEauthorrefmark{3},
        Thomas Hassan\IEEEauthorrefmark{1}, Philippe Raipin\IEEEauthorrefmark{1}
    }
    \IEEEauthorblockA{\IEEEauthorrefmark{1} Orange Innovation, Cesson-Sévigné, France}
    \IEEEauthorblockA{\IEEEauthorrefmark{2} Nantes University, École Centrale Nantes, IMT Atlantique, CNRS, INRIA, LS2N, UMR 6004, Nantes, France}
    \IEEEauthorblockA{\IEEEauthorrefmark{3} SnT, SIGCOM, University of Luxembourg, Luxembourg}
}

\maketitle

\begin{abstract}
From a telecommunication standpoint, the surge in users and services challenges next-generation networks with escalating traffic demands and limited resources. Accurate traffic prediction can offer network operators valuable insights into network conditions and suggest optimal allocation policies. Recently, spatio-temporal forecasting, employing \acp{gnn}, has emerged as a promising method for cellular traffic prediction. However, existing studies, inspired by road traffic forecasting formulations, overlook the dynamic deployment and removal of base stations, requiring the \ac{gnn}-based forecaster to handle an evolving graph. This work introduces a novel inductive learning scheme and a generalizable \ac{gnn}-based forecasting model that can process diverse graphs of cellular traffic with one-time training. We also demonstrate that this model can be easily leveraged by transfer learning with minimal effort, making it applicable to different areas. Experimental results show up to 9.8\% performance improvement compared to the state-of-the-art, especially in rare-data settings with training data reduced to below 20\%.
\end{abstract}
\acresetall
\vspace{-5pt}
\section{Introduction}
Network traffic prediction is crucial for efficient network management, providing operators with insights for optimization. In a zero-touch network, traffic prediction can continuously inform the monitoring components about future network conditions, enabling timely decision-making. Framed as a time-series forecasting problem, predicting network traffic involves identifying the best model using historical data for accurate future predictions. Recently, spatiotemporal forecasting based on \acp{gnn} has emerged as a promising approach, capturing correlations between traffic patterns of \acp{enb}. For instance, geographically proximate eNodeBs can exhibit similar traffic volumes due to comparable population densities. Additionally, depending on users' mobility patterns during network usage, network traffic state can be \textit{propagated} from one \ac{enb} to another.

Spatiotemporal forecasting typically requires a pre-constructed graph of correlations, often based on \acp{enb} proximity. Existing works~\cite{fangSDGNetHandoverAwareSpatiotemporal2022, wangSpatialTemporalCellularTraffic2022} conduct spatiotemporal prediction for an entire city in one go, relying on a single proximity graph for the city's network infrastructure. While advantageous for capturing long-range correlations, this approach becomes \textit{rigid} - heavily dependent on the input graph. Therefore, any addition or removal of \acp{enb} requires reinitialization and retraining, incurring extra computing costs. Additionally, the limited data from newly deployed \acp{enb} leads to the scarce data scenario for traffic forecasting model training. Hence, transferring the model to a different city poses challenges due to differing network infrastructures. This is also the difference between transductive and inductive graph learning where in the former, the evaluated nodes are known and in the latter, they can include even unseen nodes during training time.

To address these limitations, we propose \textbf{\textsc{Flexible}} - \textit{Forecasting Cellular Traffic by Leveraging Explicit Inductive Graph-Based Learning}. To the best of our knowledge, this is the first inductive \ac{gnn}-based model for cellular traffic prediction. Focused on forecasting individual \acp{enb}, it extracts local spatial correlation from the k-hop subgraph centered at the target \ac{enb}, combining it with temporal information for accurate predictions. Its inductive design allows operation on unseen nodes during training, ensuring adaptability. By reframing the problem to predict individual traffic within its local k-hop graph, \textsc{Flexible} homogenizes graph topologies and regularizes the spatiotemporal model, enhancing generalizability. This simplifies transfer learning into a direct scheme without additional steps, in contrast to prior methods like \cite{mallick2021transfer, huangTransferLearningTraffic2021, tang2022domain, jinTransferableGraphStructure2023, wuDeepTransferLearning2022}. Despite the limitations of exploiting only local information, \textsc{Flexbile} gains the possibility to perform prediction on unseen \acp{enb} and its straightforward transfer learning mechanism. Indeed, by experimental results, we show that when the data for training is very few, a pre-trained \textsc{Flexible} that is finetuned on a new city's traffic data can outperform state-of-the-art models.
\section{Related works}
\subsection{Graph Neural Networks - GNNs}
\Acp{gnn} are widely recognized in the deep learning literature for their ability to handle irregular data, specifically graphs. Early research focused on formulating convolutional operators for graphs, with Fast spectral filtering~\cite{cnn_graph} being a pioneering work that implemented convolution in the spectral space of graphs. Subsequent works further simplified the computation by representing the convolutional operator as a stack of 1-hop filters~\cite{gilmerNeuralMessagePassing}. This approach, often referred to as \ac{mpnn}, relies on two differentiable functions: message aggregation within a local neighborhood and node updates based on the aggregated message. In the context of spatiotemporal graphs with node features containing temporal data, leveraging \ac{gnn} insights becomes crucial for developing models adept at capturing graph dynamics.

\subsection{Spatiotemporal forecasting}
Spatiotemporal forecasting is an emerging approach to tackle cellular traffic prediction with high precision \cite{wuDeepTransferLearning2022, fangSDGNetHandoverAwareSpatiotemporal2022, wangSpatialTemporalCellularTraffic2022} and it is inspired from the road traffic forecasting \cite{li2018dcrnn_traffic, bai2020adaptive, wu2020connecting}. While extensively researched, existing methods typically address the problem in a transductive setting where the given graph remains constant during both the training and inference phases. This work proposes a more flexible model designed to operate natively in an inductive setting, accommodating scenarios where the graphs differ between the training and inference phases.

\subsection{Transfer learning}
In the domain of spatiotemporal prediction, transfer learning across different cities has been extensively explored, particularly when data is scarce. Existing works managed to find workaround solutions to handle the discrepancy between graph topologies in source and target cities. Some approaches, like graph partitioning with padding \cite{mallick2021transfer, huangTransferLearningTraffic2021}, node2vec-based feature creation \cite{tang2022domain}, clustering algorithms with inter-city region mapping \cite{wuDeepTransferLearning2022}, and the transferable mechanism for graph structure learning \cite{jinTransferableGraphStructure2023} introduce additional computing costs. The most significant drawback, however, is that all these methods necessitate re-training if the target spatial graph changes.
\section{Methodology}
\subsection{Problem definition}

\label{sect:formulation}

\begin{table}
    \centering
    \caption{Table of notation}
    \begin{tabular}{@{}ll@{}}
    \toprule
        Notation & Description \\ \midrule
        $\mX_i^{(t)}$ & Traffic value of $i$-th node at $t$-th timestep \\
        dist(i, j)  & Geographical distance between locations of i and j \\
        $T_h$ & Number of historical steps \\
        $T_f$ & Number of prediction steps \\
        $\gT$ & Set of timesteps \\
        $\gV$ & Set of \acp{enb} \\
        $\vx_{i,t}$ & $\mX_i^{(t-T_h:t)}$  \\
        $\vy_{i,t}$ & $\mX_i^{(t:t+T_f)}$  \\
        $\vh_i^{(l)}$ & Hidden features of node $i$ extracted at the $l$-th layer \\
        $p$ & Edge dropout probability \\
        $d$ & Dilation factor within the dilated convolution operator \\
        $C$ & Number of hidden features \\
        $L$ & Number of spatiotemporal blocks \\
        $\lambda$ & Weight decay \\
        $\gK$ & Set of kernel sizes \\
        $B$ & Batch size \\
        \bottomrule
    \end{tabular}
    \label{tab:notation}
\end{table}

The traffic prediction problem is defined as finding an optimal forecasting function $f$ that predicts the future $T_f$ timesteps given the historical $T_h$ timesteps. We denote the multivariate time series representing the entire traffic dataset as $\mX$ and $\mX^{(t)}_i$ as a scalar value indicating the traffic of the $i$-th variable at timestep $t$. The traffic prediction problem is defined as:
\begin{align}
    \argmin_f  \sum_{t \in \gT, i \in \gV}\gL\left(f\left(\mX_i^{(t - T_h:t)} \right), \mX_i^{(t:t + T_f)} \right),
\end{align}
where $\gL$ is the loss function, $\gT$ is the discrete temporal interval, and $\gV$ is the set of vertices - in this case - a set of \acp{enb}. Furthermore, we provide all repeatedly used notations in this paper in the \Tableref{tab:notation}.

\begin{figure*}
    \centering
    \subfloat[Transfer learning pipeline\label{1a}]{%
        \includegraphics[width=.4\textwidth]{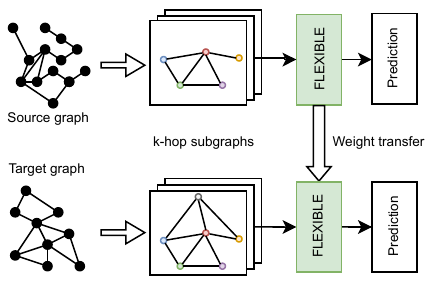}}
    \hfill
    \subfloat[Architecture of \textsc{Flexible}\label{1b}]{%
        \includegraphics[width=.55\textwidth]{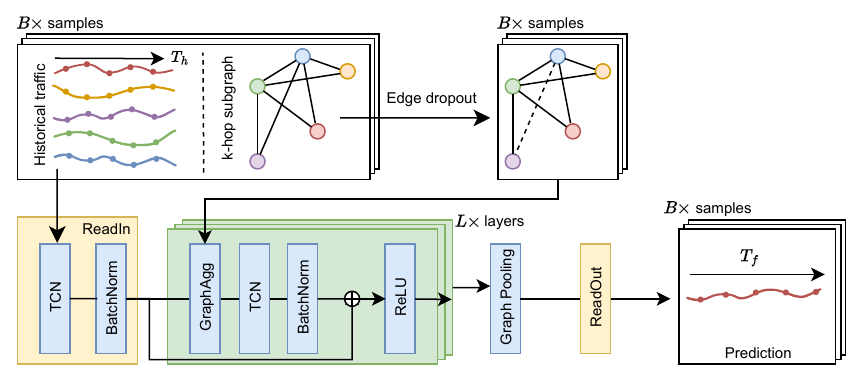}}
    \caption{The proposed frameworks: (a) transferring \textsc{Flexbile} from source to target city and (b) model architecture which takes a batch of k-hop subgraphs and associated historical traffic to generate a batch of prediction for the target \ac{enb}.}
    \label{fig:design}
\end{figure*}

In contrast to previous studies that adopt transductive settings, splitting only $\gT$, we introduce \textsc{Flexible}. This fully inductive \ac{gnn}-based model addresses the continuous deployment of eNodeBs, allowing variations in both $\gT$ and $\gV$ during different phases. To achieve this, we reframe the problem as a graph-level task instead of node-level. For predicting traffic at base station $i$ during the interval $[t, t+T_f)$, we extract the k-hop subgraph centered at node $i$. This subgraph, along with its nodes' traffic states, is fed into a learned spatiotemporal model for forecasting, represented by:
\begin{align}
\hat \vy_{i, t} = f(\mA_k(i), \{\vx_{j, t}~|~j \in \gV_k(i)\}),
\end{align}
where $\mA_k(i)$ is the adjacency matrix of the induced $k$-hop subgraph around the node $i$, $\gV_k(i)$ is set of nodes involved in the subgraph, and $f$ is the parameterized model. We visualize our pipeline in \Figref{fig:design} and describe components in the following sections.
\subsection{Graph construction}
\vspace{-3pt}
\subsubsection{Proximity graph} Since mobile users are often in movement, we can assume that the nearby \acp{enb}' traffic is correlated. Therefore, we craft a graph of correlation solely based on the geographical position of \acp{enb}:
\begin{align}
    W^s_{ij} = \begin{cases}
    \exp\left(-\textrm{dist}(i, j)\right), & \text{if $\textrm{dist}(i, j) < \kappa$ and $i \neq j$}.\\
    0, & \text{otherwise}.
  \end{cases}
\end{align}
where $\textrm{dist}(i, j)$ is the geographical distance between two \acp{enb} $i$ and $j$. In both the Paris and Lyon datasets, we set $\kappa = 3.5$km, which is the minimum distance so that both graphs remain connected. Furthermore, we sparsify the graph by limiting the maximum node degree to 10.
\vspace{-3pt}
\subsubsection{k-hop subgraphs} For each node $i$, we pre-build its k-hop subgraphs from the constructed large proximity graph and store them in a key-value database. Given a pair $(i, t)$, we can efficiently retrieve the subgraph $\mA_k(i)$ from the database. Additionally, we can access the set of traffic information linked to its vertices at time $t$ using stored node IDs associated with the subgraph. During training, we introduce Edge Dropout to these subgraphs with a probability of $p$ to augment data and mitigate overfitting.
\vspace{-5pt}
\subsection{Spatiotemporal module}
\subsubsection{Temporal Convolutional Network}
In studying time series data, different methods like \acp{cnn}, \acp{rnn}, and Attention mechanisms are used. Among all of them, the \ac{cnn}, which is natively designed for parallel computing and requires fewer parameters, is a fast and effective choice for analyzing time series data with good accuracy~\cite{bai2020adaptive, wu2020connecting, li2018dcrnn_traffic}. For our approach, a key technique is using dilated causal convolution~\cite{aaron2016wavenet}. This helps capture a broad range of information quickly while maintaining the causality in the features we extract. Furthermore, we follow the Inception-like~\cite{inception} architecture by having multiple kernel sizes in each dilated causal convolution layer to capture features at multiple scales and rapidly enlarge the receptive field. Given $\vh \in \sR^{T_h \times C}$ as the hidden features of an arbitrary node, the proposed \ac{tcn} can thus be described as below:
\begin{align}
    \textrm{TCN}(\vh) = \bigcup^{\textrm{concat}}_{K \in \gK}~\vh*_{\rm dc}\vf_K
\end{align}
where $\vf_K$ is the filter of kernel size $K$ and the resulting feature from \ac{tcn} is the concatenation of all filtered signals corresponding to every kernel size in the set $\gK$. Moreover, $*_{\rm dc}$ is the dilated causal convolution as described in~\cite{aaron2016wavenet} with the dilation exponentially scaled by a factor $d$ after each layer.
\subsubsection{Graph Isomorphism Network}
It has been demonstrated that \acp{gnn} can encounter challenges in distinguishing between different graphs in specific cases~\cite{xu2018how}, making them less powerful than the \ac{wl} graph isomorphism test. Addressing this limitation, Xu \etal~\cite{xu2018how} introduced \ac{gin}, which has been shown to possess the same discriminative power as the WL test. In the context of our forecasting problem formulated as a graph-level task, the model's ability to capture various graph topologies becomes crucial. Being able to distinguish between different graphs is essential for enhancing the model's expressive power.

Consequently, we inspire from \ac{gin} to propose our spatio-temporal model:
\begin{align}
    &\textrm{GraphAgg}(\vh_i) = (1+\epsilon)\vh_i + \sum_{u \in \gN(u)}\vh_u \label{eq:graphagg}\\
    &\vh_i^{(l+1)}            = \textrm{ReLU}\left(\textrm{TCN}(\textrm{GraphAgg}(\vh_i^{(l)})) + \vh_i^{(l)}\right)
\end{align}
where $\vh_i^{(l)} \in \sR^{T_h \times C}$ is the hidden features of an arbitrary node $i$ extracted from the $l$-th layer of the model and $\epsilon$ is a learned scalar. It has been shown in~\cite{xu2018how} that the local aggregation in \Eqref{eq:graphagg} and the \ac{tcn} followed by ReLU as an approximator are essential elements to build a \ac{gnn} as powerful as the \ac{wl} test. Additionally, this design allows the model to be \textit{flexible} so that it can process any graph topology. During transfer learning, only the scalar $\epsilon$ and \ac{tcn} weights are transferred, being independent of graph shapes. To prevent over-smoothing, a residual connection is included, and for practical regularization, batch normalization~\cite{batchnorm} is applied after each \ac{tcn}, as illustrated in \Figref{fig:design}.

\subsection{ReadIn and ReadOut}
\subsubsection{ReadIn}
Given the historical traffic $x_{i, t} \in \sR^{N \times T_h}$, we process it through a \ac{tcn} block followed by a batch normalization layer. The extracted features are further processed through a stack of $L$ spatiotemporal layers, as described above.
\subsubsection{Graph Pooling}
The final hidden features are obtained after processing the $L$-th spatiotemporal layer, resulting in $\mH \in \sR^{N \times T_h \times C}$ corresponding to hidden features of $N$ nodes in the k-hop subgraph. We then apply a graph pooling layer to pool out the final representation that should be used for the prediction. Since we only need to predict the traffic on one target node, we take out only its associated hidden features and call that TargetPooling. We also let a hyperparameter sweep to choose among this and three other poolings considered in~\cite{xu2018how}: global sum, max, and mean pooling.
\subsubsection{ReadOut}
After Graph Pooling, we acquire a hidden feature $\vh \in \sR^{T_h \times C}$. To produce the prediction, we propose a two-step ReadOut module:
\begin{align}
    \hat \vh[\tau, :] &= \textrm{ReLU}\left(\sum_{l=t-T_h}^{t-1}w_{l\tau}\vh[l,:] + a_\tau\right) \\
    \hat \vy[\tau] &= \sum_{c=0}^{C-1} z_c \hat\vh[\tau, c] + b_\tau
\end{align}
where $w_{l\tau}, a_\tau, z_c,$ and $b_\tau$ are  learnable weights.

\subsection{Loss function}
We train the model to minimize the \ac{mae} between the prediction and the ground truth while adding a regularization on the norm of the model's parameters to prevent overfitting:
\begin{align}
        \gL = \frac{1}{|T_f||\gT \times \gV|} \sum_{(i, t) \in \gT \times \gV} \left| \vy_{i, t} - \hat \vy_{i, t} \right| + \lambda \lVert\vTheta\rVert \label{eq:mae}
\end{align}
where $\vTheta$ is the model's learnable parameters and $\lambda$ is the regularization weight.

\section{Evaluation}
\subsection{Dataset}

To evaluate the performance of \textsc{Flexible}, we utilize the NetMob dataset~\cite{netmob23} which covers 77 days of the traffic demand, from March 16, 2019, to May 31, 2019,  generated by 68 mobile services across 20 metropolitan areas in France. This is so far the largest and the most recent cellular traffic dataset that we have found. For our experiments, we aggregate the downlink traffic of 5 prominent services in Paris and Lyon: Apple Video, Fortnite, Netflix, Instagram, and Microsoft Mail. These applications were chosen as they collectively represent the majority of cellular traffic. Moreover, they offer a diverse range of content consumption types, aligned with the Quality of Service Class Identifier~\cite{3gpp}. By examining the aggregated traffic of these 5 distinct applications, we can benchmark the model's capability to handle complex traffic dynamics.

However, the provided data is the traffic per $100\times100m^2$ tile which is aggregated from several \acp{enb}' traffic based on distance. This format deviates from the real-world scenario where data is typically collected and aggregated per \ac{enb}, which is essential for forecasting traffic and performing radio resource management. To address this issue, we combine the spatiotemporal data with the \ac{enb} map~\cite{cartoradio} and use Voronoi tessellation to re-aggregate the traffic into per-\ac{enb} traffic:
\begin{align}
    \textrm{Vor}(i) &= \{\textrm{tile}_{m}~|~\textrm{dist}(\textrm{tile}_{m}, i) < \textrm{dist}(\textrm{tile}_{m}, j) \forall j \in \gV \} \\
    \mX^{(t)}_i &= \sum_{m \in \textrm{Vor}(i)} \mT^{(t)}_m
\end{align}
where $\mT^{(t)}_m$ is the raw traffic of the $m$-th tile at time $t$ and Vor($i$) is the Voronoi partition of the $i$-th \ac{enb}. Finally, we obtain 1555 and 1230 \acp{enb} for Paris and Lyon respectively, and 7392 timesteps for each as the data resolution is 1 sample per 15 minutes.
\subsection{Experimental settings}\label{sec:exp_set}
\subsubsection{Overview}
Our model is initially trained on Paris traffic in an inductive setting, serving as a pre-trained model for subsequent transfer to the Lyon dataset. Hyperparameter tuning is carried out during this phase. For fine-tuning Lyon traffic, we adopt a transductive setting to facilitate comparisons with state-of-the-art models. In any setting, data is split into three parts: train, validation, and test sets. Training occurs on the train set, with model selection and hyperparameter tuning based on validation set results. Reported results for comparison come from inference on the test set, following the common setting $T_h = 12, T_f=3$, which are equivalent to 3 hours and 45 minutes respectively.

\subsubsection{Data split}

As elaborated in \Secref{sect:formulation}, every data sample can be retrieved via $(i, t) \in \gT \times \gV$. Accordingly, we split the data by splitting the set $\gT \times \gV$. We ensure that $|\gT_\textrm{train}| = 0.7 |\gT|$, $|\gT_\textrm{val}| = 0.1 |\gT|$, and $|\gT_\textrm{test}| = 0.2 |\gT|$. The same factors go for $\gV$ splitting. Consequently, the 3 data splits in the inductive setting are $\gT_\textrm{train} \times \gV_\textrm{train}$, $\gT_\textrm{val} \times \gV_\textrm{val}$, and $\gT_\textrm{test} \times \gV_\textrm{test}$. In a transductive setting, $\gV_\textrm{train} = \gV_\textrm{val} = \gV_\textrm{test} = \gV$.

\subsubsection{Evaluation metrics}
We adopt two common metrics to evaluate the models' performance. Given the prediction horizon $h$, the metrics at each horizon are defined as:
\begin{itemize}
    \item \Ac{rmse}:
    \begin{align}
        \textrm{RMSE}_h = \sqrt{\frac{1}{|\gT_\textrm{test}||\gV_\textrm{test}|} \sum_{i, t} (\vy_{i,t}[h] - \hat \vy_{i,t}[h])^2}
    \end{align}
    \item \Acf{mae}:
    \begin{align}
        \textrm{MAE}_h = \frac{1}{|\gT_\textrm{test}||\gV_\textrm{test}|} \sum_{i, t} \left\lvert\vy_{i,t}[h] - \hat \vy_{i,t}[h]\right\rvert
    \end{align}
\end{itemize}

\subsection{Hyperparameter tuning}

To tune model hyperparameters effectively, we use Optuna~\cite{optuna_2019}, a hyperparameter search framework. We employ default settings for the optimizer, utilizing the Tree-structured Parzen Estimator algorithm~\cite{NIPS2011_86e8f7ab} for continual downsampling of the search space. Additionally, the median stopping rule is applied for early pruning of less promising trials. The search space and final hyperparameter values are summarized in \Tableref{tab:hyper_tune}.
\begin{table}[]
\caption{Hyperparameters' search space and their optimal values~\cite{optuna_2019}.}
\centering
\begin{tabular}{@{}lcc@{}}
\toprule
Hyperparameters & Search space                         & Optimal value     \\ \midrule
Learning rate   & 0.001, 0.003, 0.005, ..., 0.019      & 0.009     \\
$p$             & 0, 0.05, 0.1, 0.15, 0.2              & 0.05      \\
$d$             & 1, 2, 3                              & 1         \\
$C$             & 32, 64, 128, 256                     & 64        \\
$L$             & 1, 2, 3, 4, 5                        & 2         \\
$\lambda$       & $10^{-4}, 10^{-5}, 10^{-6}, 10^{-7}$ & $10^{-5}$ \\
$\gK$           & \{1, 3\}, \{3\}, \{1, 3, 5, 7\}      & \{1, 3\}  \\
$B$             & \{256, 512, 1024, 2048, 4096\}       & 4096      \\
Graph Pooling   & Mean, Max, Sum, Target                    & Target    \\ \bottomrule
\end{tabular}
\label{tab:hyper_tune}
\end{table}
\vspace{-3pt}
\subsection{Baselines}
\subsubsection{Univariate models}
\begin{itemize}
    \item \newterm{LSTM}~\cite{lstm} is a sophisticated neural network architecture meticulously crafted for the nuanced task of capturing and retaining information within long sequential data.
    \item \newterm{TCN} is a variant of \textsc{Flexible}, where all graph-related components are removed from the architecture.
\end{itemize}
\subsubsection{Multivariate models}
\begin{itemize}
    \item \textbf{DCRNN}~\cite{li2018dcrnn_traffic} is a spatiotemporal model, which leverages dual directional diffusion convolution to capture spatial dependencies and incorporates it into a sequence of Gated Recurrent Units to further exploit the temporal dimension.
    \item \textbf{AGCRN}~\cite{bai2020adaptive} integrates adaptive graph convolutional networks within a sequence of \acp{gru} to capture both node-specific spatial and temporal correlations in traffic series.
    \item \textbf{MTGNN}~\cite{wu2020connecting} incorporates self-adaptive graph convolution with mix-hop propagation layers for spatial modules and dilated inception layers for temporal modules.
\end{itemize}

\subsection{Results of inductive learning}
This experiment simulates a real-world scenario where a model trained on traffic data of existing \acp{enb} is used to predict traffic on unseen \acp{enb}. Since multivariate baselines are not natively adapted to inductive settings, in this experiment, we only compare our method with univariate models. Conducted on Paris cellular traffic data, the results in \Tableref{tab:res_ind} showcase \textsc{Flexible} outperforming other models across all prediction horizons and evaluation metrics. This underscores the advantage of leveraging spatial correlation and \textsc{Flexible}'s ability to handle graph topology discrepancy between training and inference. Moreover, the comparable performance between \textsc{LSTM} and \textsc{TCN} indicates \textsc{TCN}'s effectiveness in capturing sequential information akin to \textsc{LSTM}.
\begin{table}[]
\caption{Results of inductive learning on the cellular traffic in Paris ($\times 10^6$). The results are averaged over 3 runs for each model.}
\label{tab:res_ind}
\centering
\begin{tabular}{@{}ccccc@{}}
\toprule
Metrics & Horizon & LSTM & TCN & FLEXIBLE \\ \midrule
\multirow{3}{*}{MAE} & 15 min & 2.70 \scriptsize{\tpm 0.004} & 2.71 \scriptsize{\tpm 0.007} & \textbf{2.62} \scriptsize{\tpm 0.003} \\
 & 30 min & 3.19 \scriptsize{\tpm 0.002} & 3.21 \scriptsize{\tpm 0.005} & \textbf{3.04} \scriptsize{\tpm 0.002} \\
 & 45 min & 3.55 \scriptsize{\tpm 0.004} & 3.56 \scriptsize{\tpm 0.011} & \textbf{3.34} \scriptsize{\tpm 0.012 }\\ \midrule
\multirow{3}{*}{RMSE} & 15 min & 4.53 \scriptsize{\tpm 0.016} & 4.54 \scriptsize{\tpm 0.019} & \textbf{4.42} \scriptsize{\tpm 0.021} \\
 & 30 min & 5.26 \scriptsize{\tpm 0.023} & 5.26 \scriptsize{\tpm 0.010} & \textbf{5.05} \scriptsize{\tpm 0.021} \\
 & 45 min & 5.64 \scriptsize{\tpm 0.026} & 5.63 \scriptsize{\tpm 0.008} & \textbf{5.35} \scriptsize{\tpm 0.005} \\ \bottomrule
\end{tabular}
\end{table}

\subsection{Results of transductive learning with full data}
To evaluate the effectiveness of our approach relative to multivariate models, we conduct experiments in transductive settings as outlined in Section \ref{sec:exp_set}. Additionally, we leverage the inductive capability of \textsc{Flexible} for transfer learning across cities. In this specific trial, we explore traffic data in Lyon, concurrently fine-tuning the \textsc{Flexible} model previously trained on Paris data to enhance its forecasting performance on Lyon data. We call the transferred model \textsc{Tr-Flexible.}

The results of each model when they are trained or finetuned on 100\% of the data, which is 66 days of traffic, are summarized in \Tableref{tab:res_tra_100}. In this setting, MTGNN stays coherent with its claim to outperform other existing methods. Our model, \textsc{flexbile}, due to its \textit{local} nature, cannot achieve prediction as precise as state-of-the-art models. It also highlights that transfer learning does improve the forecasting precision, enabling \textsc{Tr-Flexible} to perform comparably to AGCRN, even better at RMSE-15min and RMSE-30min. However, we believe that these shortcomings become minor when training data is fewer, allowing \textsc{Tr-Flexbile} to be a better alternative to \textit{global} models when forecasting newly deployed \acp{enb}. In the next section, we present the quantitative results of the data scarcity scenario and demonstrate the performance improvement of \textsc{Flexible} when data is reduced to below 20\%.
\begin{table}[]
\centering
\caption{Results of transductive learning on the cellular traffic in Lyon ($\times10^5$).}
\label{tab:res_tra_100}
\begin{tabular}{@{}lcccccc@{}}
\toprule
Metrics & \multicolumn{3}{c}{MAE} & \multicolumn{3}{c}{RMSE} \\ \cmidrule(lr){2-4} \cmidrule(l){5-7}
Horizon (min) & 15      & 30      & 45     & 15      & 30      & 45      \\ \midrule
AGCRN & 5.04   & 5.63   & 6.05  & 9.73   & 10.69  & 11.31  \\
DCRNN & \underline{4.84} & \underline{5.38} & \underline{5.66} & \underline{9.34} & \underline{10.34} & \underline{10.80} \\
MTGNN & \textbf{4.74} & \textbf{5.19} & \textbf{5.45} & \textbf{9.19} & \textbf{10.15} & \textbf{10.62} \\
FLEXIBLE   & 5.13   & 5.97   & 6.47  & 9.84   & 10.92  & 11.72  \\ \midrule
TR-FLEXIBLE & 4.98 & 5.81 & 6.36 & 9.41 & 10.51 & 11.41 \\ \bottomrule
\end{tabular}
\end{table}

\subsection{Results of data-scarcity setting}
To assess the forecasting capabilities of newly deployed \acp{enb}, we systematically reduce the volume of training-validation data, investigating the impact on model performance under varying levels of data scarcity: 5\%, 10\%, 20\%, and 40\%. In essence, we maintain the test set $\gT_{\rm test}$ unchanged to ensure fair evaluation while progressively dropping the oldest time steps in the training data. This approach ensures that the models are consistently trained with the most recent data, allowing for a comprehensive analysis of their adaptability to limited training information. The validation split is included in this training-validation set and its proportion to the training split is maintained at 1/7. The results are reported in \Tableref{tab:res_few}.

\begin{figure}
    \centering
    \includegraphics[width=.7\linewidth]{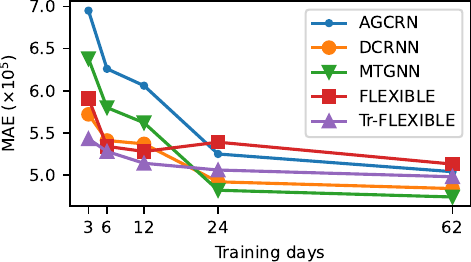}
    \caption{The MAE at 15min-horizon for the amount of data used for training.}
    \label{fig:mae-wrt-days}
\end{figure}

We observe that the performance of all models degrades as training data becomes more limited. While, at 40\% of data, the performance gaps between models do not change significantly compared to the full data setting. However starting from 20\% down to 5\%, \textsc{Tr-Flexible} takes the lead in terms of prediction precision. Notably, in the 10\% and 20\% scenarios, \textsc{Flexible}, trained exclusively on the target dataset, also outperforms baseline models. This could be due to the \textit{global} models struggling to capture neither temporal nor spatial long-range dependency with limited training data. Moreover, with the complicated design and numerous parameters, the baselines are easily overfitted when the training data is scarce. In contrast, our model, with a compact size and a focus on short-range dependencies, proves more suitable for such situations. We also illustrate the variation of \ac{mae}-15min of all models for the amount of training data in \Figref{fig:mae-wrt-days}. Overall, these results highlight the robustness of our model in scenarios with limited data, suitable for traffic prediction of new \acp{enb} and knowledge transfer between cities.
\begin{table*}[]
\centering
\caption{\ac{mae} of each model in few-shot learning on the cellular traffic in Lyon ($\times10^5$).}
\label{tab:res_few}
\begin{tabular}{@{}lccccccccccccr@{}}
\toprule
\multicolumn{1}{l}{Scarcity rate} & \multicolumn{3}{c}{5\% ($\sim$3 days)} & \multicolumn{3}{c}{10\% ($\sim$6 days)} & \multicolumn{3}{c}{20\% ($\sim$12 days)} & \multicolumn{3}{c}{40\% ($\sim$24 days)} & Number of\\ \cmidrule(lr){2-4} \cmidrule(lr){5-7} \cmidrule(lr){8-10} \cmidrule(l){11-13}
\multicolumn{1}{l}{Horizon (min)}  & 15 & 30 & 45 & 15 & 30 & 45 & 15 & 30 & 45 & 15 & 30 & 45 & parameters \\ \midrule
 AGCRN & 6.95 & 8.24 & 9.14 & 6.26 & 7.50 & 8.34 & 6.06 & 7.24 & 8.20 & 5.25 & 5.93 & 6.45 & 1514715\\
 DCRNN & \underline{5.72} & 6.67 & \underline{7.36} & 5.41 & 6.61 & 7.56 & 5.37 & 6.54 & 7.35 & \underline{4.92} & \underline{5.55} & \underline{5.96} & 372353\\
 MTGNN & 6.38 & 7.47 & 8.44 & 5.80 & 6.96 & 8.25 & 5.62 & 6.62 & 7.75 & \textbf{4.82} & \textbf{5.33} & \textbf{5.65} & 1861043\\
 FLEXIBLE & 5.91 & \underline{6.61} & 7.50 & \underline{5.34} & \underline{6.30} & \underline{7.37} & \underline{5.28} & \underline{6.10} & \underline{6.82} & 5.39 & 6.16 & 6.81 & 140970\\ \midrule
 Tr-FLEXIBLE & \textbf{5.43} & \textbf{6.10} & \textbf{6.64} & \textbf{5.28} & \textbf{6.20} & \textbf{7.03} & \textbf{5.14} & \textbf{5.99} & \textbf{6.67} & 5.06 & 5.92 & 6.49 & 140970\\
 \bottomrule
\end{tabular}
\end{table*}

\section{Conclusion}
In this paper, we propose \textsc{Flexible}, a novel inductive graph-based model for cellular traffic forecasting. We first formulate the problem and describe in detail the dataset we are working on and the technical details of the proposed model. Finally, we conduct experiments in 3 settings: inductive learning on Paris data, transfer learning with full data in Lyon, and transfer learning with few data in Lyon. The experimental results prove the capability of learning in inductive and demonstrate the power of \textsc{Flexbile} in data-scarcity situations, suggesting it as an alternative model for newly deployed \acp{enb}. Future research should focus on enhancing the model's expressivity for improved prediction accuracy and exploring continual learning techniques for evolving \ac{enb} deployment scenarios.
\section*{Acknowledgement}
This research was performed using data made available by Orange within the NetMob 2023 Data Challenge~\cite{netmob23}.

\bibliographystyle{IEEEtran}
\bibliography{citation}

\end{document}

%% file: acronyms.tex
\begin{acronym}
\acro{gin}[GIN]{Graph Isomorphism Network}
\acro{wl}[WL]{Weisfeiler-Lehman}
\acro{lstm}[LSTM]{Long Short-Term Memory}
\acro{gnn}[GNN]{Graph Neural Network}
\acro{tcn}[TCN]{Temporal Convolutional Network}
\acro{cnn}[CNN]{Convolutional Neural Network}
\acro{rnn}[RNN]{Recurrent Neural Network}
\acro{mpnn}[MPNN]{Message-Passing Neural Network}
\acro{rmse}[RMSE]{Root Mean Square Error}
\acro{mae}[MAE]{Mean Absolute Arror}
\acro{dtw}[DTW]{Dynamic Time Warping}
\acro{enb}[eNB]{Evolved NodeB}
\acro{gru}[GRU]{Gated Recurrent Unit}
\end{acronym}

%% file: commands.tex
\def\etal{\textit{et al.}}

\usepackage{amsmath,amsfonts,bm}

\newcommand{\tpm}{$\pm~$}

\newcommand{\newterm}[1]{{\bf #1}}

\def\Tableref#1{Table~\ref{#1}}

\def\Figref#1{Figure~\ref{#1}}

\def\Secref#1{Section~\ref{#1}}

\def\eqref#1{equation~\ref{#1}}
\def\Eqref#1{Equation~\ref{#1}}

\def\1{\bm{1}}

\def\vTheta{{\bm{\Theta}}}

\def\vf{{\bm{f}}}

\def\vh{{\bm{h}}}

\def\vx{{\bm{x}}}
\def\vy{{\bm{y}}}

\def\mA{{\bm{A}}}

\def\mH{{\bm{H}}}

\def\mT{{\bm{T}}}

\def\mX{{\bm{X}}}

\DeclareMathAlphabet{\mathsfit}{\encodingdefault}{\sfdefault}{m}{sl}
\SetMathAlphabet{\mathsfit}{bold}{\encodingdefault}{\sfdefault}{bx}{n}

\def\gK{{\mathcal{K}}}
\def\gL{{\mathcal{L}}}

\def\gN{{\mathcal{N}}}

\def\gT{{\mathcal{T}}}

\def\gV{{\mathcal{V}}}

\def\sR{{\mathbb{R}}}

\DeclareMathOperator*{\argmin}{arg\,min}

%% file: conference_101719.bbl
\begin{thebibliography}{10}
\providecommand{\url}[1]{#1}
\csname url@samestyle\endcsname
\providecommand{\newblock}{\relax}
\providecommand{\bibinfo}[2]{#2}
\providecommand{\BIBentrySTDinterwordspacing}{\spaceskip=0pt\relax}
\providecommand{\BIBentryALTinterwordstretchfactor}{4}
\providecommand{\BIBentryALTinterwordspacing}{\spaceskip=\fontdimen2\font plus
\BIBentryALTinterwordstretchfactor\fontdimen3\font minus
  \fontdimen4\font\relax}
\providecommand{\BIBforeignlanguage}[2]{{%
\expandafter\ifx\csname l@#1\endcsname\relax
\typeout{** WARNING: IEEEtran.bst: No hyphenation pattern has been}%
\typeout{** loaded for the language `#1'. Using the pattern for}%
\typeout{** the default language instead.}%
\else
\language=\csname l@#1\endcsname
\fi
#2}}
\providecommand{\BIBdecl}{\relax}
\BIBdecl

\bibitem{fangSDGNetHandoverAwareSpatiotemporal2022}
Y.~Fang, S.~Ergut, and P.~Patras, ``{{SDGNet}}: {{A Handover-Aware
  Spatiotemporal Graph Neural Network}} for {{Mobile Traffic Forecasting}},''
  \emph{IEEE Communications Letters}, vol.~26, no.~3, pp. 582--586, Mar. 2022.

\bibitem{wangSpatialTemporalCellularTraffic2022}
Z.~Wang, J.~Hu, G.~Min, Z.~Zhao, Z.~Chang, and Z.~Wang, ``Spatial-{{Temporal
  Cellular Traffic Prediction}} for 5 {{G}} and {{Beyond}}: {{A Graph Neural
  Networks-Based Approach}},'' \emph{IEEE Transactions on Industrial
  Informatics}, pp. 1--10, 2022.

\bibitem{mallick2021transfer}
T.~Mallick, P.~Balaprakash, E.~Rask, and J.~Macfarlane, ``Transfer learning
  with graph neural networks for short-term highway traffic forecasting,'' in
  \emph{2020 25th International Conference on Pattern Recognition
  (ICPR)}.\hskip 1em plus 0.5em minus 0.4em\relax IEEE, 2021, pp.
  10\,367--10\,374.

\bibitem{huangTransferLearningTraffic2021}
Y.~Huang, X.~Song, S.~Zhang, and J.~J. Yu, ``Transfer {{Learning}} in {{Traffic
  Prediction}} with {{Graph Neural Networks}},'' in \emph{2021 {{IEEE
  International Intelligent Transportation Systems Conference}}
  ({{ITSC}})}.\hskip 1em plus 0.5em minus 0.4em\relax {IEEE}, Sep. 2021, pp.
  3732--3737.

\bibitem{tang2022domain}
Y.~Tang, A.~Qu, A.~H. Chow, W.~H. Lam, S.~Wong, and W.~Ma, ``Domain adversarial
  spatial-temporal network: A transferable framework for short-term traffic
  forecasting across cities,'' in \emph{Proceedings of the 31st ACM
  International Conference on Information \& Knowledge Management}, 2022, pp.
  1905--1915.

\bibitem{jinTransferableGraphStructure2023}
Y.~Jin, K.~Chen, and Q.~Yang, ``Transferable {{Graph Structure Learning}} for
  {{Graph-based Traffic Forecasting Across Cities}},'' in \emph{Proceedings of
  the 29th {{ACM SIGKDD Conference}} on {{Knowledge Discovery}} and {{Data
  Mining}}}.\hskip 1em plus 0.5em minus 0.4em\relax {ACM}, Aug. 2023, pp.
  1032--1043.

\bibitem{wuDeepTransferLearning2022}
Q.~Wu, K.~He, X.~Chen, S.~Yu, and J.~Zhang, ``Deep {{Transfer Learning Across
  Cities}} for {{Mobile Traffic Prediction}},'' \emph{IEEE/ACM Transactions on
  Networking}, vol.~30, no.~3, pp. 1255--1267, Jun. 2022.

\bibitem{cnn_graph}
M.~Defferrard, X.~Bresson, and P.~Vandergheynst, ``Convolutional neural
  networks on graphs with fast localized spectral filtering,'' in
  \emph{Advances in Neural Information Processing Systems}, 2016.

\bibitem{gilmerNeuralMessagePassing}
J.~Gilmer, S.~S. Schoenholz, P.~F. Riley, O.~Vinyals, and G.~E. Dahl, ``Neural
  message passing for quantum chemistry,'' in \emph{Proceedings of the 34th
  International Conference on Machine Learning - Volume 70}, ser.
  ICML'17.\hskip 1em plus 0.5em minus 0.4em\relax JMLR.org, 2017, p.
  1263–1272.

\bibitem{li2018dcrnn_traffic}
Y.~Li, R.~Yu, C.~Shahabi, and Y.~Liu, ``Diffusion convolutional recurrent
  neural network: Data-driven traffic forecasting,'' in \emph{International
  Conference on Learning Representations (ICLR '18)}, 2018.

\bibitem{bai2020adaptive}
L.~Bai, L.~Yao, C.~Li, X.~Wang, and C.~Wang, ``Adaptive graph convolutional
  recurrent network for traffic forecasting,'' \emph{Advances in neural
  information processing systems}, vol.~33, pp. 17\,804--17\,815, 2020.

\bibitem{wu2020connecting}
Z.~Wu, S.~Pan, G.~Long, J.~Jiang, X.~Chang, and C.~Zhang, ``Connecting the
  dots: Multivariate time series forecasting with graph neural networks,'' in
  \emph{Proceedings of the 26th ACM SIGKDD international conference on
  knowledge discovery \& data mining}, 2020, pp. 753--763.

\bibitem{aaron2016wavenet}
A.~van~den Oord, S.~Dieleman, H.~Zen, K.~Simonyan, O.~Vinyals, A.~Graves,
  N.~Kalchbrenner, A.~W. Senior, and K.~Kavukcuoglu, ``Wavenet: {A} generative
  model for raw audio,'' \emph{CoRR}, vol. abs/1609.03499, 2016.

\bibitem{inception}
C.~Szegedy, W.~Liu, Y.~Jia, P.~Sermanet, S.~Reed, D.~Anguelov, D.~Erhan,
  V.~Vanhoucke, and A.~Rabinovich, ``Going deeper with convolutions,'' in
  \emph{2015 IEEE Conference on Computer Vision and Pattern Recognition
  (CVPR)}, 2015, pp. 1--9.

\bibitem{xu2018how}
K.~Xu, W.~Hu, J.~Leskovec, and S.~Jegelka, ``How powerful are graph neural
  networks?'' in \emph{International Conference on Learning Representations},
  2019.

\bibitem{batchnorm}
S.~Ioffe and C.~Szegedy, ``Batch normalization: accelerating deep network
  training by reducing internal covariate shift,'' in \emph{Proceedings of the
  32nd International Conference on International Conference on Machine Learning
  - Volume 37}, ser. ICML'15.\hskip 1em plus 0.5em minus 0.4em\relax JMLR.org,
  2015, p. 448–456.

\bibitem{netmob23}
O.~E. Mart{\'{\i}}nez{-}Durive, S.~Mishra, C.~Ziemlicki, S.~Rubrichi,
  Z.~Smoreda, and M.~Fiore, ``The netmob23 dataset: {A} high-resolution
  multi-region service-level mobile data traffic cartography,'' \emph{CoRR},
  vol. abs/2305.06933, 2023.

\bibitem{3gpp}
``{3GPP TS 23.203 Policy and Charging Control Architecture V17.2.0},'' Dec.
  2021.

\bibitem{cartoradio}
T.~N.~F. Agency, ``Cartoradio: The map of radio sites and wave measurements.''

\bibitem{optuna_2019}
T.~Akiba, S.~Sano, T.~Yanase, T.~Ohta, and M.~Koyama, ``Optuna: A
  next-generation hyperparameter optimization framework,'' in \emph{Proceedings
  of the 25th {ACM} {SIGKDD} International Conference on Knowledge Discovery
  and Data Mining}, 2019.

\bibitem{NIPS2011_86e8f7ab}
J.~Bergstra, R.~Bardenet, Y.~Bengio, and B.~K\'{e}gl, ``Algorithms for
  hyper-parameter optimization,'' in \emph{Advances in Neural Information
  Processing Systems}, J.~Shawe-Taylor, R.~Zemel, P.~Bartlett, F.~Pereira, and
  K.~Weinberger, Eds., vol.~24.\hskip 1em plus 0.5em minus 0.4em\relax Curran
  Associates, Inc., 2011.

\bibitem{lstm}
S.~Hochreiter and J.~Schmidhuber, ``Long short-term memory,'' \emph{Neural
  Comput.}, vol.~9, no.~8, p. 1735–1780, nov 1997.

\end{thebibliography}
